%% file: ms.tex
\documentclass[11pt,letterpaper]{article}
\usepackage{emnlp2017}
\usepackage{times}
\usepackage{latexsym}
\usepackage[utf8]{inputenc}
\input{macro}

\usepackage{booktabs} 
\usepackage{subfigure, multirow, tabularx}
\usepackage[boxed, ruled,vlined]{algorithm2e}
\usepackage{booktabs} 

\emnlpfinalcopy

\def\emnlppaperid{1321}

\title{Heterogeneous Supervision for Relation Extraction: \\A Representation Learning Approach}

\author{
Liyuan Liu$^{\dag *}\ $ 
Xiang Ren$^{\dag}\thanks{\enspace Equal contribution.}\quad$ Qi Zhu$^{\dag}\quad$ Huan Gui$^{\dag}\quad$ Shi Zhi$^{\dag}\quad$ Heng Ji$^{\sharp}\quad$ Jiawei Han$^{\dag}$\\[0.5ex]
{$^{\dag}$ University of Illinois at Urbana-Champaign, Urbana, IL, USA}\\
{$^{\sharp}$ Computer Science Department, Rensselaer Polytechnic Institute, USA}\\
{
\begin{small}
$^{\dag}$\{ll2, xren7, qiz3, huangui2, shizhi2, hanj\}@illinois.edu $^{\sharp}$\{jih\}@rpi.edu
\end{small}
}
}

\begin{document}

\maketitle

\input{abs}

\begin{figure*}[ht!]
  \centering
    \includegraphics[width=\textwidth]{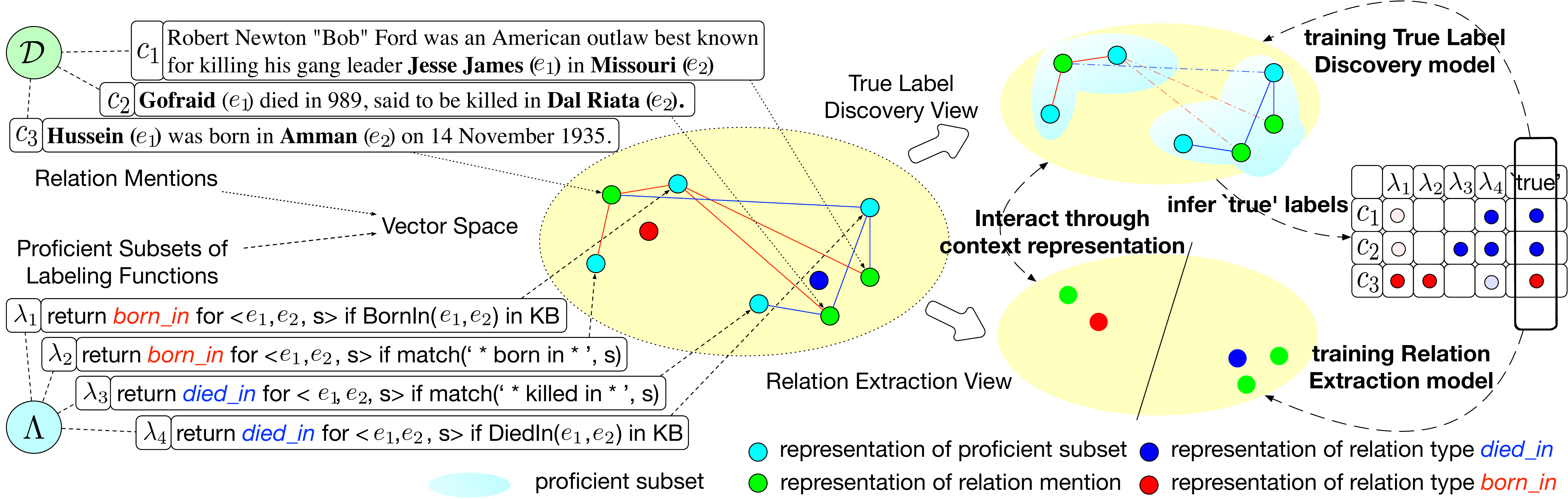}
  \caption{\our Framework except Extraction and Representation of Text Features}
  \label{fig:Framework}
\end{figure*}

\input{intro}
\begin{figure}[t]
  \centering
  \centerline{
    \includegraphics[width=\columnwidth]{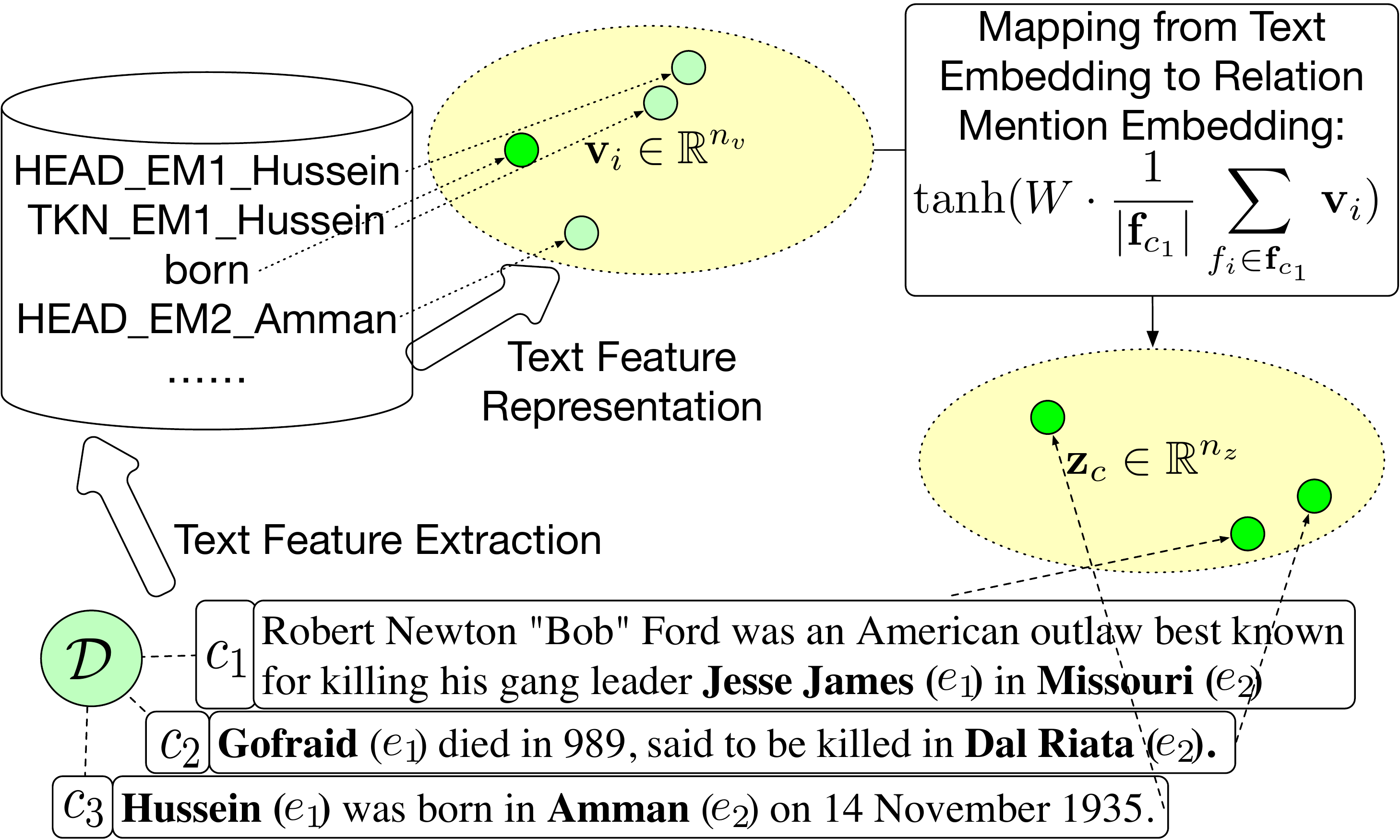}
  }
  \caption{Relation Mention Representation}
  \label{fig:context_rep}
\end{figure}

\input{prelim}
\input{method}
\input{exp}
\input{related}
\input{conclusion}
\input{acknowledge}

\newpage
\bibliography{cited}
\bibliographystyle{emnlp_natbib}

\end{document}

%% file: macro.tex
\usepackage{amsmath, amsfonts, amssymb, xspace, color,listings,enumitem,graphicx}

\newcommand{\hide}[1]{} 

\def \E {\mathbb{E}}

\def \f {\mathbf{f}}

\def \l {\mathbf{l}}

\def \t {\mathbf{t}}

\def \v {\mathbf{v}}

\def \z {\mathbf{z}}
\def \A {\mathbf{A}}

\def \C {\mathcal{C}}
\def \D {\mathcal{D}}
\def \F {\mathcal{F}}

\def \O {\mathcal{O}}

\def \R {\mathcal{R}}
\def \S {\mathcal{S}}

\def \ob {o}

\newlength\inputlen

\newlength\outputlen

\newcommand{\our}{\textsc{REHession}\xspace}
\newcommand{\ourtd}{\textsc{REHession-TD}\xspace}
\newcommand{\ourus}{\textsc{REHession-US}\xspace}
\newcommand{\ds}{\emph{distant supervision}\xspace}

\newcommand{\hs}{\emph{heterogeneous supervision}\xspace}
\newcommand{\lf}{\emph{labeling functions}\xspace}
\newcommand{\sca}{{source consistency assumption}\xspace}

\DeclareMathOperator*{\argmax}{argmax} 



%% file: abs.tex
\begin{abstract}
Relation extraction is a fundamental task in information extraction.
Most existing methods have heavy reliance on annotations labeled by human experts, which are costly and time-consuming.
To overcome this drawback, we propose a novel framework, \our, to conduct relation extractor learning using annotations from heterogeneous information source, e.g., knowledge base and domain heuristics.
These annotations, referred as \hs, often conflict with each other, which brings a new challenge
 to the original relation extraction task: how to infer the true label from noisy labels for a given instance.
Identifying context information as the backbone of both relation extraction and true label discovery, we adopt embedding techniques to learn the distributed representations of context, which bridges all components with mutual enhancement in an iterative fashion.
Extensive experimental results demonstrate the superiority of \our over the state-of-the-art.
\end{abstract} 

%% file: intro.tex
\section{Introduction}
\label{sect:intro}
One of the most important tasks towards text understanding is to detect and categorize semantic relations between two entities in a given context.
For example, in Fig.~\ref{fig:Framework}, with regard to the sentence of $c_1$, relation between \emph{Jesse James} and \emph{Missouri} should be categorized as \texttt{died\_in}. 
With accurate identification, relation extraction systems can provide essential support for many applications.
One example is question answering, regarding a specific question, relation among entities can provide valuable information, which helps to seek better answers~\cite{bao2014knowledge}.
Similarly, for medical science literature, relations like protein-protein interactions~\cite{fundel2007relex} and gene disease associations~\cite{chun2006extraction} can be extracted and used in knowledge base population.
Additionally, relation extractors can be used in ontology construction~\cite{schutz2005relext}.


Typically, existing methods follow the supervised learning paradigm, and require extensive annotations from domain experts, which are costly and time-consuming.
To alleviate such drawback, attempts have been made to build relation extractors with a small set of seed instances or human-crafted patterns~\cite{nakashole2011scalable,carlson2010coupled}, 
based on which more patterns and instances will be iteratively generated by bootstrap learning.
However, these methods often suffer from semantic drift~\cite{mintz2009distant}.
Besides, knowledge bases like Freebase have been leveraged to automatically generate training data and provide \ds\cite{mintz2009distant}.
Nevertheless, for many domain-specific applications, distant supervision is either non-existent or insufficient (usually less than $25\%$ of relation mentions are covered~\cite{ren2015clustype,ling2012fine}).


Only recently have preliminary studies been developed to unite different supervisions, including knowledge bases and domain specific patterns, which are referred as \hs.
As shown in Fig. 1, these supervisions often conflict with each other~\cite{ratner2016data}.
To address these conflicts, data programming~\cite{ratner2016data} employs a generative model, which encodes supervisions as \lf, and adopts the \sca: \emph{a source is likely to provide true information with the same probability for all instances}.
This assumption is widely used in true label discovery literature~\cite{Li:2016:STD:2897350.2897352} to model reliabilities of information sources like crowdsourcing and infer the true label from noisy labels. Accordingly, most true label discovery methods would trust a human annotator on all instances to the same level.

However, labeling functions, unlike human annotators, do not make casual mistakes but follow certain ``error routine''. 
Thus, the reliability of a labeling function is not consistent among different pieces of instances.
In particular, a labeling function could be more reliable for a certain subset~\cite{varma2016socratic} (also known as its \emph{proficient subset}) comparing to the rest. 
We identify these proficient subsets based on context information, only trust labeling functions on these subsets and avoid assuming global source consistency.

Meanwhile, embedding methods have demonstrated great potential in capturing semantic meanings, which also reduce the dimension of overwhelming text features.
Here, we present \our, a novel framework capturing context's semantic meaning through representation learning, and conduct both relation extraction and true label discovery in a context-aware manner.
Specifically, as depicted in Fig.~\ref{fig:Framework}, we embed relation mentions in a low-dimension vector space, where similar relation mentions tend to have similar relation types and annotations. 
`True' labels are further inferred based on reliabilities of labeling functions, which are calculated with their proficient subsets' representations.
Then, these inferred true labels would serve as supervision for all components, including context representation, true label discovery and relation extraction.
Besides, the context representation bridges relation extraction with true label discovery, and allows them to enhance each other.

To the best of our knowledge, the framework proposed here is the first method that utilizes representation learning to provide heterogeneous supervision for relation extraction.
The high-quality context representations serve as the backbone of true label discovery and relation extraction. 
Extensive experiments on benchmark datasets demonstrate significant improvements over the state-of-the-art.

The remaining of this paper is organized as follows. 
Section~\ref{sect:prelim} gives the definition of relation extraction with heterogeneous supervision. 
We then present the \our model and the learning algorithm in Section~\ref{sect:method}, and report our experimental evaluation in Section~\ref{sect:exp}.
Finally, we briefly survey related work in Section~\ref{sect:related} and conclude this study in Section~\ref{sect:conclusion}. 

%% file: prelim.tex

\section{Preliminaries}
\label{sect:prelim}

In this section, we would formally define relation extraction and heterogeneous supervision, including the format of labeling functions.

\subsection{Relation Extraction}
\label{subsec:re}
Here we conduct relation extraction in \textit{sentence-level}~\cite{bao2014knowledge}.
For a sentence $d$, an entity mention is a token span in $d$ which represents an entity, and a relation mention is a triple $(e_1, e_2, d)$ which consists of an ordered entity pair $(e_1, e_2)$ and $d$. And the relation extraction task is to categorize relation mentions into a given set of relation types $\R$, or Not-Target-Type (\texttt{None}) which means the type of the relation mention does not belong to $\R$.

\subsection{Heterogeneous Supervision}
\label{subsec:hetersup}
Similar to \cite{ratner2016data}, we employ labeling functions as basic units to encode supervision information and generate annotations. Since different supervision information may have different proficient subsets, we require each labeling function to encode only one elementary supervision information. Specifically, in the relation extraction scenario, we require each labeling function to only annotate one relation type based on one elementary piece of information, e.g., four examples are listed in Fig.~\ref{fig:Framework}.

Notice that knowledge-based labeling functions are also considered to be noisy because relation extraction is conducted in sentence-level, e.g. although \texttt{president\_of} (\emph{Obama}, \emph{USA}) exists in KB, it should not be assigned with ``\emph{Obama} was born in Honolulu, Hawaii, \emph{USA}'', since \texttt{president\_of} is irrelevant to the context.


\subsection{Problem Definition}
\label{subsec:pd}
For a POS-tagged corpus $\D$ with detected entities, we refer its relation mentions as $\C = \{c_i = (e_{i,1}, e_{i,2}, d), \forall d \in \D\}$. 
Our goal is to annotate entity mentions with relation types of interest ($\R = \{r_1, \dots, r_K\}$) or \texttt{None}. 
We require users to provide heterogeneous supervision in the form of labeling function $\Lambda = \{\lambda_1, \dots, \lambda_M\}$, and mark the annotations generated by $\Lambda$ as $\O = \{o_{c, i} | \lambda_i \text{ generate annotation }o_{c, i} \text{ for }c \in \C\}$. 
We record relation mentions annotated by $\Lambda$ as $\C_l$, and refer relation mentions without annotation as $\C_u$. 
Then, our task is to train a relation extractor based on $\C_l$ and categorize relation mentions in $\C_u$.

%% file: method.tex


\section{The \our Framework}
\label{sect:method}

\input{notation}

Here, we present \our, a novel framework to infer true labels from automatically generated noisy labels, and categorize unlabeled instances into a set of relation types.
Intuitively, errors of annotations ($\O$) come from  mismatch of contexts, e.g., in Fig.~\ref{fig:Framework}, $\lambda_1$ annotates $c_1$ and $c_2$ with 'true' labels but for mismatched contexts `killing' and 'killed'. 
Accordingly, we should only trust labeling functions on \emph{matched} context, e.g., trust $\lambda_1$ on $c_3$ due to its context `was born in', but not on $c_1$ and $c_2$. 
On the other hand, relation extraction can be viewed as \emph{matching} appropriate relation type to a certain context.
These two \emph{matching} processes are closely related and can enhance each other, while context representation plays an important role in both of them.

\smallskip
\noindent
\textsf{\textbf{\small Framework Overview. }}
We propose a general framework to learn the relation extractor from automatically generated noisy labels. 
As plotted in Fig.~\ref{fig:Framework}, distributed representation of context bridges relation extraction with true label discovery, and allows them to enhance each other.
Specifically, it follows the steps below:
\begin{enumerate}[fullwidth,itemindent=0em,label=\arabic*.]\setlength{\itemsep}{-0.05cm}
\item
After being extracted from context, text features are embedded in a low dimension space by representation learning (see Fig.~\ref{fig:context_rep});

\item
Text feature embeddings are utilized to calculate relation mention embeddings (see Fig.~\ref{fig:context_rep});

\item
With relation mention embeddings, true labels are inferred by calculating labeling functions' reliabilities in a context-aware manner (see Fig.~\ref{fig:Framework});

\item
Inferred true labels would `supervise' all components to learn model parameters (see Fig.~\ref{fig:Framework}).
\end{enumerate}
We now proceed by introducing these components of the model in further details. 






\input{feature_table}

\subsection{Modeling Relation Mention }
As shown in Table~\ref{table:features}, we extract abundant lexical features \cite{ren2016cotype,mintz2009distant} to characterize relation mentions. 
However, this abundance also results in the gigantic dimension of original text features ($\sim10^7$ in our case). 
In order to achieve better generalization ability, we represent relation mentions with low dimensional ($\sim10^2$) vectors. 
In Fig.~\ref{fig:context_rep}, for example, relation mention $c_3$ is first represented as bag-of-features. 
After learning text feature embeddings, we use the average of feature embedding vectors to derive the embedding vector for $c_3$.


\smallskip
\noindent
\textsf{\textbf{\small Text Feature Representation. }} 
Similar to other principles of embedding learning, we assume text features occurring in the same contexts tend to have similar meanings (also known as distributional hypothesis\cite{harris1954distributional}). Furthermore, we let each text feature's embedding vector to predict other text features occurred in the same relation mentions or context. Thus, text features with similar meaning should have similar
 embedding vectors. Formally, we mark text features as $\F = \{f_1, \cdots, f_{|\F|}\}$, record the feature set for $\forall c \in \C$ as $\f_c$, and represent the embedding vector for $f_i$ as $\v_i \in \mathbb{R}^{n_v}$, and we aim to maximize the following log likelihood: {\small $\sum_{c \in \C_l} \sum_{f_i, f_j \in \f_c} \log \: p(f_i | f_j)$}, where {\small $p(f_i | f_j) = \exp(\v_i^T\v_j^*) / \sum_{f_k \in \F} \exp(\v_i^T \v_k^*)$}.

However, the optimization of this likelihood is impractical because the calculation of $\nabla p(f_i|f_j)$ requires summation over all text features, whose size exceeds $10^7$ in our case. 
In order to perform efficient optimization, we adopt the negative sampling technique \cite{mikolov2013distributed} to avoid this summation. 
Accordingly, we replace the log likelihood with Eq.~\ref{eqn:neg_sampling} as below:

\begin{small}
\begin{equation}
\mathcal{J}_E = \sum_{\substack{c \in \C_l \\ f_i,f_j\in \f_c}} (\log \: \sigma(\v_i^T \v_j^*) - \sum_{k=1}^V \E_{f_{k'} \sim \hat{P}} [\log \: \sigma(-\v_i^T \v_{k'}^*)]) \label{eqn:neg_sampling}
\end{equation}
\end{small}
where $\hat{P}$ is noise distribution used in~\cite{mikolov2013distributed}, $\sigma$ is the sigmoid function and $V$ is number of negative samples.

\smallskip
\noindent
\textsf{\textbf{\small Relation Mention Representation.}} 
With text feature embeddings learned by Eq.~\ref{eqn:neg_sampling}, a naive method to represent relation mentions is to concatenate or average its text feature embeddings.
However, text features embedding may be in a different semantic space with relation types.
Thus, we directly learn a mapping $g$ from text feature representations to relation mention representations  \cite{van2016learning,van2016unsupervised} instead of simple heuristic rules like concatenate or average (see Fig.~\ref{fig:context_rep}):

\begin{small}
\begin{equation}
\z_c = g(\f_c)= \text{tanh}(W \cdot \frac{1}{|\f_c|} \sum_{f_i \in \f_c} \v_i) \label{eqn:activate}
\end{equation}
\end{small}
where  $\z_c$ is the representation of $c \in \C_l$, $W$ is a $n_z \times n_v$ matrix, $n_z$ is the dimension of relation mention embeddings and tanh is the element-wise hyperbolic tangent function.

In other words, we represent bag of text features with their average embedding,
then apply linear map and hyperbolic tangent to transform the embedding from text feature semantic space to relation mention semantic space.
The non-linear tanh function allows non-linear class boundaries in other components, and also regularize relation mention representation to range $[-1, 1]$ which avoids numerical instability issues.

\subsection{True Label Discovery }
\begin{figure}[t]
  \centering
  \centerline{
    \includegraphics[width=0.8\columnwidth]{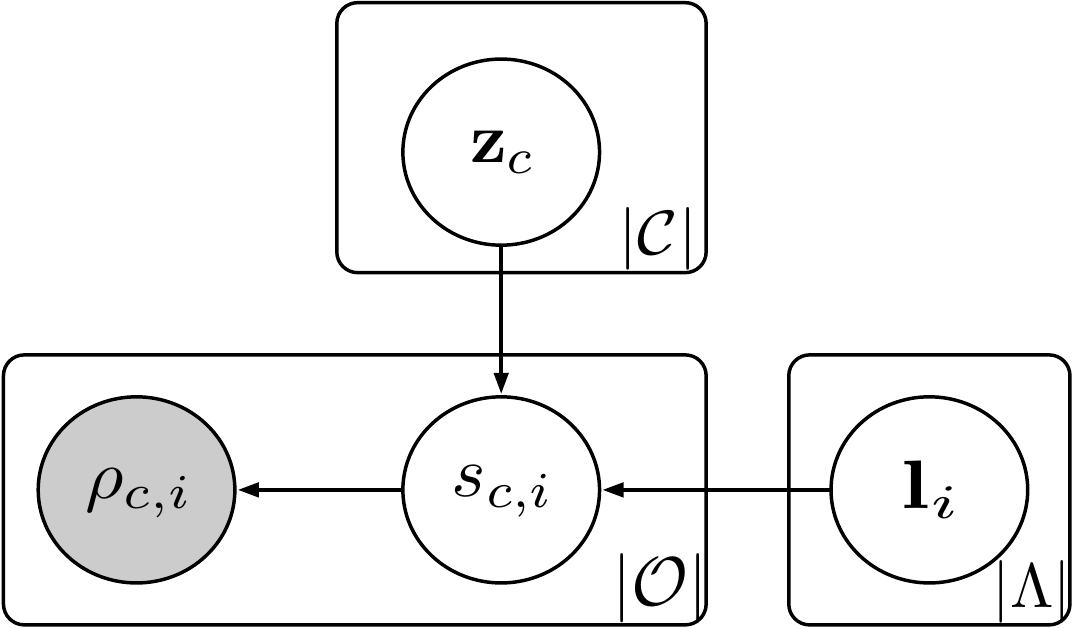}
  }
  \caption{Graphical model of $o_{c, i}$'s correctness}
  \label{fig:graphicalModel}
\end{figure}

Because heterogeneous supervision generates labels in a discriminative way, we suppose its errors follow certain underlying principles, i.e., if a labeling function annotates a instance correctly / wrongly, it would annotate other similar instances correctly / wrongly. 
For example, $\lambda_1$ in Fig.~\ref{fig:Framework} generates wrong annotations for two similar instances $c_1$, $c_2$ and would make the same errors on other similar instances.
Since context representation captures the semantic meaning of relation mention and would be used to identify relation types, we also use it to identify the mismatch of context and labeling functions. 
Thus, we suppose for each labeling function $\lambda_i$, there exists an proficient subset $\S_i$ on $\R^{n_z}$, containing instances that $\lambda_i$ can precisely annotate. 
In Fig.~\ref{fig:Framework}, for instance, $c_3$ is in the proficient subset of $\lambda_1$, while $c_1$ and $c_2$ are not.
Moreover, the generation of annotations are not really random, and we propose a probabilistic model to describe the level of mismatch from labeling functions to real relation types instead of annotations' generation.


As shown in Fig.~\ref{fig:graphicalModel}, we assume the indicator of whether $c$ belongs to $\S_i$, $s_{c, i} = \delta(c \in \S_i)$, would first be generated based on context representation 
\begin{align}
p(s_{c, i} = 1 | \z_c, \l_i) = p(c \in \S_i) = \sigma(\z_c^T \l_i) \label{eqn:gene_d}
\end{align}

Then the correctness of annotation $\ob_{c, i}$, $\rho_{c, i} = \delta(\ob_{c, i} = \ob_c^*)$, would be generated. Furthermore, we assume $p(\rho_{c, i} = 1 | s_{c, i} = 1) = \phi_1$ and $p(\rho_{c, i} = 1 | s_{c, i} = 0) = \phi_0$ to be constant for all relation mentions and labeling functions.

Because $s_{c, i}$ would not be used in other components of our framework, we integrate out $s_{c, i}$ and write the log likelihood as

\begin{small}
\begin{align}
&\mathcal{J}_T =\sum_{\ob_{c, i} \in \O} \log(\sigma(\z_c^T \l_i)\phi_1^{ \delta(\ob_{c, i} = \ob_c^*)}(1-\phi_1)^{\delta(\ob_{c, i} \neq \ob_c^*)}\nonumber\\
&+(1-\sigma(\z_c^T \l_i))\phi_0^{\delta(\ob_{c, i} = \ob_c^*)}(1-\phi_0)^{\delta(\ob_{c, i} \neq \ob_c^*)}) \label{eqn:truth_ciscovery}
\end{align}
\end{small}

Note that $\ob_c^*$ is a hidden variable but not a model parameter, and $\mathcal{J}_T$ is the likelihood of $\rho_{c, i} = \delta(\ob_{c, i} = \ob_c^*)$. 
Thus, we would first infer $\ob_c^* = \argmax_{\ob_c^*} \mathcal{J}_T$, then train the true label discovery model by maximizing $\mathcal{J}_T$.

\begin{table}[t]
\centering
\begin{tabular}{r|c|c}
\hline
\textbf{Datasets} & \textbf{NYT} & \textbf{Wiki-KBP} \\
\hline
\hline
\% of \texttt{None} in Training & 0.6717 & 0.5552\\
\hline
\% of \texttt{None} in Test & 0.8972 & 0.8532\\
\hline
\end{tabular}
\caption{Proportion of \texttt{None} in Training/Test Set}
\label{tab:twoKLF}
\end{table}

\subsection{Modeling Relation Type }
We now discuss the model for identifying relation types based on context representation. For each relation mention $c$, its representation $\z_c$ implies its relation type, and the distribution of relation type can be described by the soft-max function:

\begin{small}
\begin{equation}
p(r_i | \z_c) = \frac{\exp(\z_c^T\t_i)}{\sum_{r_j \in \R\cup\{\texttt{None}\}} \exp(\z_c^T \t_j)} \label{eqn:rel_softmax}
\end{equation}
\end{small}
where $\t_i \in \mathbb{R}^{v_z}$ is the representation for relation type $r_i$. Moreover, with the inferred true label $\ob_c^*$, the relation extraction model can be trained as a multi-class classifier. Specifically, we use Eq.~\ref{eqn:rel_softmax} to approach the distribution

\begin{small}
\begin{equation}
p(r_i |\ob_c^*) = \left\{
\begin{aligned}
1 && r_i = \ob_c^* \\
0 && r_i \neq \ob_c^*
\end{aligned}
\right.
\label{eqn:one_hot_dist}
\end{equation}
\end{small}

Moreover, we use KL-divergence to measure the dissimilarity between two distributions, and formulate model learning as maximizing $\mathcal{J}_R$:

\begin{small}
\begin{align}
\mathcal{J}_R &= - \sum_{c \in \C_l} KL(p(. | \z_c)|| p(. |\ob_c^*)) \label{eqn:kl-divergence}
\end{align}
\end{small}
where {\small $KL(p(. | \z_c)|| p(. |\ob_c^*))$} is the KL-divergence from {\small  $p(r_i |\ob_c^*)$ } to {\small $p(r_i | \z_c)$}, {\small $p(r_i | \z_c)$} and {\small $p(r_i |\ob_c^*)$} has the form of Eq.~\ref{eqn:rel_softmax} and Eq.~\ref{eqn:one_hot_dist}.

\subsection{Model Learning}

Based on Eq.~\ref{eqn:neg_sampling}, Eq.~\ref{eqn:truth_ciscovery} and Eq.~\ref{eqn:kl-divergence}, we form the joint optimization problem for model parameters as

\begin{small}
\begin{align}
&\min_{W, \v, \v^*, \l, \t, \ob^*} \mathcal{J} = -\mathcal{J}_R - \lambda_1 \mathcal{J}_E - \lambda_2 \mathcal{J}_T \nonumber\\
&\text{s.t. } \forall c \in \C_l, \ob_c^* = \argmax_{\ob_c^*} \mathcal{J}_T, \z_c = g(\f_c) \label{eqn:joint}
\end{align}
\end{small}
Collectively optimizing Eq.~\ref{eqn:joint} allows heterogeneous supervision guiding all three components, while these components would refine the context representation, and enhance each other.


In order to solve the joint optimization problem in Eq.~\ref{eqn:joint} efficiently, we adopt the stochastic gradient descent algorithm to update $\{W, \v, \v^*, \l, \t\}$ iteratively, and $\ob_c*$ is estimated by maximizing $\mathcal{J}_T$ after calculating $\z_c$. Additionally, we apply the widely used dropout techniques~\cite{srivastava2014dropout} to prevent overfitting and improve generalization performance. 

The learning process of \our is summarized as below. In each iteration, we would sample a relation mention $c$ from $\C_l$, then sample $c$'s text features and conduct the text features' representation learning. After calculating the representation of $c$, we would infer its true label $\ob_c^*$ based on our true label discovery model, and finally update model parameters based on $\ob_c^*$.


\subsection{Relation Type Inference}
\label{subsec:type_infer}
We now discuss the strategy of performing type inference for $\C_u$.
As shown in Table~\ref{tab:twoKLF}, the proportion of \texttt{None} in $\C_u$ is usually much larger than in $\C_l$. 
Additionally, not like other relation types in $\R$, \texttt{None} does not have a coherent semantic meaning. Similar to~\cite{ren2016cotype}, we introduce a heuristic rule: identifying a relation mention as \texttt{None} when (1) our relation extractor predict it as \texttt{None}, or (2) the entropy of {\small $p(. | \z_c)$} over $\R$ exceeds a pre-defined threshold $\eta$. 
The entropy is calculated as {\small $H(p(. | \z_c)) = - \sum_{r_i \in \R} p(r_i | \z_c) log (p(r_i | \z_c))$}.
And the second situation means based on relation extractor this relation mention is not likely belonging to any relation types in $\R$. 


%% file: notation.tex


\begin{table}[t]
\begin{small}
\begin{tabular}{r|l}
\hline
$\f_c$	&	$c$'s text features set, where $c \in \C$\\
$\v_i$	&	text feature embedding  for $f_i \in \F$\\
$\z_c$	&	relation mention embedding  for $c \in \C$\\
$\l_i$	&	embedding for $\lambda_i$'s proficient subset, $\lambda_i \in \Lambda$\\
$o_{c, i}$ &	annotation for $c$, generated by labeling function $\lambda_i$ \\
$o_c^*$	&	underlying true label for $c$ \\

$\rho_{c, i}$ & identify whether $o_{c, i}$ is correct \\
$\S_i$	&	the proficient subset of labeling function $\lambda_i$\\
$s_{c, i}$ & identify whether $c$ belongs to $\lambda_i$'s proficient subset\\
$\t_i$	&	relation type embedding for $r_i \in \R$\\
\hline
\end{tabular}
\end{small}
\caption{Notation Table.}
\label{tab:notation}
\end{table}


%% file: feature_table.tex

\begin{table*}
\begin{center}
\begin{small}
\begin{tabular}{lll}
\hline
\textbf{Feature} & \textbf{Description} & \textbf{Example} \\
\hline
Entity mention (EM) head & Syntactic head token of each entity mention & ``\textit{HEAD\_EM1\_Hussein}", ...
\\ 
Entity Mention Token & Tokens in each entity mention & ``\textit{TKN\_EM1\_Hussein}", ...
\\ 
Tokens between two EMs & Tokens between two EMs & ``\textit{was}", ``\textit{born}", ``\textit{in}"
\\
Part-of-speech (POS) tag & POS tags of tokens between two EMs & ``\textit{VBD}", ``\textit{VBN}", ``\textit{IN}"
\\ 
Collocations & Bigrams in left/right 3-word window of each EM & ``\textit{Hussein was}", ``\textit{in Amman}" 
\\ 
Entity mention order & Whether EM 1 is before EM 2 & ``\textit{EM1\_BEFORE\_EM2}"\\ 
Entity mention distance & Number of tokens between the two EMs & ``\textit{EM\_DISTANCE\_3}" \\ 
Body entity mentions numbers& Number of EMs between the two EMs & ``\textit{EM\_NUMBER\_0}" \\ 
Entity mention context & Unigrams before and after each EM & ``\textit{EM\_AFTER\_was}", ... \\ 
Brown cluster (learned on $\D$) & Brown cluster ID for each token  & ``\textit{BROWN\_010011001}", ... \\ 
\hline
\end{tabular}
\end{small}
\caption{ \small Text features $\F$ used in this paper. (``\textit{Hussein}'', ``\textit{Amman}'',``\textit{\textbf{Hussein} was born in \textbf{Amman}}") is used as an example.}
\label{table:features}
\end{center}
\end{table*}

%% file: exp.tex

\section{Experiments}
\label{sect:exp}

In this section, we empirically validate our method by comparing to the state-of-the-art relation extraction methods on news and Wikipedia articles.


\subsection{Datasets and settings}
In the experiments, we conduct investigations on two benchmark datasets from different domains:\footnote{ Codes and datasets used in this paper can be downloaded at: \url{https://github.com/LiyuanLucasLiu/ReHession}.}

\noindent
\textbf{\small NYT} \cite{riedel2010modeling} is a news corpus sampled from $\sim$ 294k 1989-2007 New York Times news articles. It consists of 1.18M sentences, while 395 of them are annotated by authors of \cite{hoffmann2011knowledge} and used as test data;

\noindent
\textbf{\small Wiki-KBP} utilizes 1.5M sentences sampled from 780k Wikipedia articles \cite{ling2012fine} as training corpus, while test set consists of the 2k sentences manually annotated in 2013 KBP slot filling assessment results \cite{ellis2012linguistic}.

For both datasets, the training and test sets partitions are maintained in our experiments. 
Furthermore, we create validation sets by randomly sampling $10\%$ mentions from each test set and used the remaining part as evaluation sets. 

\smallskip
\noindent
\textsf{\small\textbf{Feature Generation. }} 
As summarized in Table~\ref{table:features}, we use a 6-word window to extract context features for each entity mention, 
apply the Stanford CoreNLP tool \cite{manning2014stanford} to generate entity mentions and get POS tags for both datasets. Brown clusters\cite{brown1992class} are derived for each corpus using public implementation\footnote{\url{https://github.com/percyliang/brown-cluster}}. All these features are shared with all compared methods in our experiments.


\begin{table}[t]
\centering
\begin{footnotesize}
\begin{tabular}{r||c|c||c|c}
\hline
\multirow{2}{*}{Kind}& \multicolumn{2}{c||}{Wiki-KBP} & \multicolumn{2}{c}{NYT} \\
\cline{2-5}
 & \#Types & \#LF & \#Types & \#LF\\
\hline
Pattern& 13 & 147 & 16 & 115\\
\hline
KB& 7 & 7 & 25 & 26\\
\hline
\end{tabular}
\end{footnotesize}
\caption{\small Number of labeling functions and the relation types they can annotated w.r.t. two kinds of information}
\label{tab:lf_stats}
\end{table}


\smallskip
\noindent
\textsf{\small\textbf{Labeling Functions. }} 
In our experiments, labeling functions are employed to encode two kinds of supervision information. 
One is knowledge base, the other is handcrafted domain-specific patterns. 
For domain-specific patterns, we manually design a number of labeling functions\footnote{pattern-based labeling functions can be accessed at: \url{https://github.com/LiyuanLucasLiu/ReHession}};
for knowledge base, annotations are generated following the procedure in~\cite{ren2016cotype,riedel2010modeling}.

Regarding two kinds of supervision information, the statistics of the labeling functions are summarized in Table~\ref{tab:lf_stats}.
We can observe that heuristic patterns can identify more relation types for KBP datasets, while for NYT datasets, knowledge base can provide supervision for more relation types.
This observation aligns with our intuition that single kind of information might be insufficient while different kinds of information can complement each other.

We further summarize the statistics of annotations in Table~\ref{tab:data_conflicts}.
It can be observed that a large portion of instances is only annotated as \texttt{None}, while lots of conflicts exist among other instances. This phenomenon justifies the motivation to employ true label discovery model to resolve the conflicts among supervision. Also, we can observe most conflicts involve \texttt{None} type, accordingly, our proposed method should have more advantages over traditional true label discovery methods on the relation extraction task comparing to the relation classification task that excludes \texttt{None} type.

\input{relationExtraction}

\begin{table}[t]
\centering
\begin{small}
\begin{tabular}{l|c|c}
\hline
\textbf{Dataset} & \textbf{Wiki-KBP} & \textbf{NYT} \\
\hline
Total Number of RM& 225977 & 530767\\
\hline
RM annotated as \texttt{None} & 100521  & 356497\\
\hline
RM with conflicts & 32008 & 58198\\
\hline
Conflicts involving \texttt{None} & 30559 & 38756 \\
\hline
\end{tabular}
\end{small}
\caption{\small Number of relation mentions (RM), relation mentions annotated as \texttt{None}, relation mentions with conflicting annotations and conflicts involving \texttt{None}}
\label{tab:data_conflicts}
\end{table}

\subsection{Compared Methods} 
We compare \our with below methods: 

\noindent
\textbf{\small FIGER}~\cite{ling2012fine} adopts multi-label learning with Perceptron algorithm.

\noindent
\textbf{\small BFK}~\cite{bunescu2005subsequence} applies bag-of-feature kernel to train a support vector machine;

\noindent
\textbf{\small DSL}~\cite{mintz2009distant} trains a multi-class logistic classifier\footnote{We use liblinear package from \url{https//github.com/cjlin1/liblinear}} on the training data;

\noindent
\textbf{\small MultiR}~\cite{hoffmann2011knowledge} models training label noise by multi-instance multi-label learning;

\noindent
\textbf{\small FCM}~\cite{gormley2015improved} performs compositional embedding by neural language model.

\noindent
\textbf{\small CoType-RM}~\cite{ren2016cotype} adopts partial-label loss to handle label noise and train the extractor.

Moreover, two different strategies are adopted to feed heterogeneous supervision to these methods. The first is to keep all noisy labels, marked as `NL'. 
Alternatively, a true label discovery method, Investment~\cite{pasternack2010knowing}, is applied to resolve conflicts, which is based on the source consistency assumption and iteratively updates inferred true labels and label functions' reliabilities. 
Then, the second strategy is to only feed the inferred true labels, referred as `TD'.

Universal Schemas~\cite{riedel2013relation} is proposed to unify different information by calculating a low-rank approximation of the annotations $\O$. 
It can serve as an alternative of the Investment method, i.e., selecting the relation type with highest score in the low-rank approximation as the true type. 
But it doesn’t explicitly model noise and not fit our scenario very well. 
Due to the constraint of space, we only compared our method to Investment in most experiments, and Universal Schemas is listed as a baseline in Sec.~\ref{subsec:case}. 
Indeed, it performs similarly to the Investment method. 

\smallskip
\noindent
\textsf{\small\textbf{Evaluation Metrics. }} 
For relation classification task, which excludes \texttt{None} type from training / testing, we use the classification accuracy (Acc) for evaluation, and for relation extraction task, precision (Prec), recall (Rec) and F1 score~\cite{bunescu2005subsequence,bach2007review} are employed.
Notice that both relation extraction and relation classification are conducted and evaluated in \textit{sentence-level}~\cite{bao2014knowledge}.

\smallskip
\noindent
\textsf{\small\textbf{Parameter Settings. }} 
Based on the semantic meaning of proficient subset, we set $\phi_2$ to {\small $1 / |\R \cup \{\texttt{None}\}|$}, i.e., the probability of generating right label with random guess. 
Then we set $\phi_1$ to $1 - \phi_2$, $\lambda_1 = \lambda_2 = 1$, and the learning rate $\alpha = 0.025$. 
As for other parameters, they are tuned on the validation sets for each dataset. 
Similarly, all parameters of compared methods are tuned on validation set, and the parameters achieving highest F1 score are chosen for relation extraction.

\begin{table}
\centering
\begin{scriptsize}
\begin{tabular}{p{3.3 cm}|c|c}
\hline
\multirow{2}{*}{\textbf{Relation Mention}} & \multirow{2}{*}{\textbf{\our}} & \textbf{Investment \&} \\
& & \textbf{Universal Schemas}\\
\hline
\hline
\textit{Ann Demeulemeester} ( \textbf{born} 1959 , Waregem , \textit{Belgium} ) is ... & \texttt{born-in} & \texttt{None} \\
\hline
\textit{Raila Odinga} was \textbf{born} at ..., in \textit{Maseno}, Kisumu District, ... & \texttt{born-in} & \texttt{None}\\
\hline
\hline
\textit{Ann Demeulemeester} ( \textbf{elected} 1959 , Waregem , \textit{Belgium} ) is ... & \texttt{None} & \texttt{None} \\
\hline
\textit{Raila Odinga} was \textbf{examined} at ..., in \textit{Maseno}, Kisumu District, ... & \texttt{None} & \texttt{None} \\
\hline
\end{tabular}
\end{scriptsize}
\caption{\small Example output of true label discovery. The first two relation mentions come from Wiki-KBP, and their annotations are \{\texttt{born-in}, \texttt{None}\}. The last two are created by replacing key words of the first two. Key words are marked as bold and entity mentions are marked as Italics.}
\label{tab:case_study}
\end{table}

\subsection{Performance Comparison}

Given the experimental setup described above, the averaged evaluation scores in 10 runs of relation classification and relation extraction on two datasets are summarized in Table~\ref{table:relation_extraction}.

From the comparison, it shows that NL strategy yields better performance than TD strategy, since the true labels inferred by Investment are actually wrong for many instances. 
On the other hand, as discussed in Sec.~\ref{subsec:case}, our method introduces context-awareness to true label discovery, while the inferred true label guides the relation extractor achieving the best performance. 
This observation justifies the motivation of avoiding the source consistency assumption and the effectiveness of proposed true label discovery model.

One could also observe the difference between \our and the compared methods is more significant on the NYT dataset than on the Wiki-KBP dataset. This observation accords with the fact that the NYT dataset contains more conflicts than KBP dataset (see Table~\ref{tab:data_conflicts}), and the intuition is that our method would have more advantages on more conflicting labels.


Among four tasks, the relation classification of Wiki-KBP dataset has highest label quality, i.e. conflicting label ratio, but with least number of training instances.
And CoType-RM and DSL reach relatively better performance among all compared methods.
CoType-RM performs much better than DSL on Wiki-KBP relation classification task, while DSL gets better or similar performance with CoType-RM on other tasks. 
This may be because the representation learning method is able to generalize better, thus performs better when the training set size is small. 
However, it is rather vulnerable to the noisy labels compared to DSL. 
Our method employs embedding techniques, and also integrates context-aware true label discovery to de-noise labels, making the embedding method rather robust, thus achieves the best performance on all tasks.





\subsection{Case Study}
\label{subsec:case}


\smallskip
\noindent
\textsf{\small\textbf{Context Awareness of True Label Discovery. }} 

Although Universal Schemas does not adopted the source consistency assumption, but it's conducted in document-level, and is context-agnostic in our sentence-level setting. Similarly, most true label discovery methods adopt the source consistency assumption, which means if they trust a labeling function, they would trust it on all annotations. And our method infers true labels in a context-aware manner, which means we only trust labeling functions on matched contexts.

For example, Investment and Universal Schemas refer \texttt{None} as true type for all four instances in Table~\ref{tab:case_study}. And our method infers \texttt{born-in} as the true label for the first two relation mentions; after replacing the matched contexts (\textbf{born}) with other words (\textbf{elected} and \textbf{examined}), our method no longer trusts \texttt{born-in} since the modified contexts are no longer matched, then infers \texttt{None} as the true label. In other words, our proposed method infer the true label in a context aware manner.

\smallskip
\noindent
\textsf{\small\textbf{Effectiveness of True Label Discovery. }} 
We explore the effectiveness of the proposed context-aware true label discovery component by comparing \our to its variants \ourtd and \ourus, which uses Investment or Universal Schemas to resolve conflicts. The averaged evaluation scores are summarized in Table~\ref{tab:context-awareTD}. 
We can observe that \our significantly outperforms its variants. 
Since the only difference between \our and its variants is the model employed to resolve conflicts, this gap verifies the effectiveness of the proposed context-aware true label discovery method.

\begin{table}[t]
\centering
\begin{small}
\begin{tabular}{l|l|c|c|c|c}
\hline
\multicolumn{2}{c|}{\textbf{\scriptsize Dataset \& Method}} &\textbf{Prec} &\textbf{Rec} & \textbf{F1} & \textbf{Acc} \\
\hline
\hline
\multirow{3}{*}{\textbf{\scriptsize Wiki-KBP}} 
& Ori & \textbf{0.3677} & 0.4933 & \textbf{0.4208} & \textbf{0.7277} \\
\cline{2-6}
& TD & 0.3032 & \textbf{0.5279} & 0.3850 & 0.7271\\
\cline{2-6}
& US & 0.3380 & 0.4779 & 0.3960 & 0.7268\\
\hline
\multirow{3}{*}{\textbf{\scriptsize NYT}} & Ori & \textbf{0.4122} & \textbf{0.5726} & \textbf{0.4792} & \textbf{0.8381} \\
\cline{2-6}
& TD & 0.3758 & 0.4887 & 0.4239 & 0.7387 \\
\cline{2-6}
& US & 0.3573 & 0.5145 & 0.4223 & 0.7362\\
\hline
\end{tabular}
\end{small}
\caption{\small Comparison among \our (Ori), \ourus(US) and \ourtd (TD) on relation extraction and relation classification}
\label{tab:context-awareTD}
\end{table}







%% file: relationExtraction.tex

\begin{table*}
\begin{center}
\begin{tabular}{  l || c|c|c | c|c|c || c|c}
\hline
\multirow{3}{*}{\textbf{Method}} & \multicolumn{6}{c||}{\textbf{Relation Extraction}} & \multicolumn{2}{c}{\textbf{Relation Classification}} \\
\cline{2-9}
& \multicolumn{3}{c|}{\textbf{NYT}} &  \multicolumn{3}{c||}{\textbf{Wiki-KBP}} & \textbf{NYT} & \textbf{Wiki-KBP}\\
\cline{2-9}
& \textbf{Prec} & \textbf{Rec}  & \textbf{F1}
& \textbf{Prec} & \textbf{Rec} & \textbf{F1}
& \textbf{Accuracy} & \textbf{Accuracy}
\\ \hline
NL+FIGER
& 0.2364 & 0.2914 & 0.2606  
& 0.2048 & 0.4489 & 0.2810 
& 0.6598 & 0.6226 
\\
NL+BFK
& 0.1520 & 0.0508 & 0.0749 
& 0.1504 & 0.3543 & 0.2101 
& 0.6905 & 0.5000   
\\
NL+DSL
& 0.4150 & 0.5414 & 0.4690 
& 0.3301 & 0.5446 & 0.4067 
& 0.7954 & 0.6355   
\\ 
NL+MultiR
& 0.5196 & 0.2755 & 0.3594 
& 0.3012 & 0.5296 & 0.3804 
& 0.7059 & 0.6484 
\\
NL+FCM
& 0.4170 & 0.2890 & 0.3414 
& 0.2523 & 0.5258 & 0.3410 
& 0.7033 & 0.5419 
\\
NL+CoType-RM
& 0.3967 & 0.4049 & 0.3977  
& \textbf{0.3701} & 0.4767 & 0.4122
& 0.6485 & 0.6935
\\
\hline
TD+FIGER
& 0.3664 & 0.3350 & 0.3495 
& 0.2650 & \textbf{0.5666} & 0.3582 
& 0.7059 & 0.6355
\\
TD+BFK
& 0.1011 & 0.0504 & 0.0670 
& 0.1432 & 0.1935 & 0.1646 
& 0.6292 & 0.5032   
\\
TD+DSL
& 0.3704 & 0.5025 & 0.4257 
& 0.2950 & 0.5757 & 0.3849 
& 0.7570 & 0.6452
\\ 
TD+MultiR
& \textbf{0.5232} & 0.2736 & 0.3586 
& 0.3045 & 0.5277 & 0.3810 
& 0.6061 & 0.6613  
\\
TD+FCM
& 0.3394 & 0.3325 & 0.3360 
& 0.1964 & 0.5645 & 0.2914
& 0.6803 & 0.5645   
\\
TD+CoType-RM
& 0.4516 & 0.3499 & 0.3923 
& 0.3107 & 0.5368 & 0.3879 
& 0.6409 & 0.6890   
\\
\hline
\our
& 0.4122 & \textbf{0.5726} & \textbf{0.4792} 
& 0.3677 & 0.4933 & \textbf{0.4208}
& \textbf{0.8381}  & \textbf{0.7277}
\\
\hline
\end{tabular}
\end{center}
\caption{Performance comparison of relation extraction and relation classification}
\label{table:relation_extraction}
\end{table*}

%% file: related.tex
\section{Related Work}
\label{sect:related}


\subsection{Relation Extraction}
Relation extraction aims to detect and categorize semantic relations between a pair of entities. To alleviate the dependency of annotations given by human experts, weak supervision~\cite{bunescu2007learning,etzioni2004web} and distant supervision~\cite{ren2016cotype} have been employed to automatically generate annotations based on knowledge base (or seed patterns/instances). 
Universal Schemas~\cite{riedel2013relation,verga2015multilingual,toutanova2015representing} has been proposed to unify patterns and knowledge base, but it's designed for document-level relation extraction, i.e., not to categorize relation types based on a specific context, but based on the whole corpus. 
Thus, it allows one relation mention to have multiple true relation types; and does not fit our scenario very well, which is sentence-level relation extraction and assumes one instance has only one relation type. 
Here we propose a more general framework to consolidate heterogeneous information and further refine the true label from noisy labels, which gives the relation extractor potential to detect more types of relations in a more precise way.

Word embedding has demonstrated great potential in capturing semantic meaning~\cite{mikolov2013distributed}, and achieved great success in a wide range of NLP tasks like relation extraction~\cite{zeng2014relation,takase2016composing,nguyen2015combining}.
In our model, we employed the embedding techniques to represent context information, and reduce the dimension of text features, which allows our model to generalize better.

\subsection{Truth Label Discovery}
True label discovery methods have been developed to resolve conflicts among multi-source information under the assumption of source consistency \cite{Li:2016:STD:2897350.2897352,zhi2015modeling}. Specifically, in the \emph{spammer-hammer} model \cite{karger2011iterative}, each source could either be a spammer, which annotates instances randomly; or a hammer, which annotates instances precisely. In this paper, we assume each labeling function would be a hammer on its proficient subset, and would be a spammer otherwise, while the proficient subsets are identified in the embedding space. 

Besides data programming, socratic learning \cite{varma2016socratic} has been developed to conduct binary classification under heterogeneous supervision. Its true label discovery module supervises the discriminative module in label level, while the discriminative module influences the true label discovery module by selecting a feature subset. Although delicately designed, it fails to make full use of the connection between these modules, i.e., not refine the context representation for classifier. Thus, its discriminative module might suffer from the overwhelming size of text features. 

%% file: conclusion.tex
\section{Conclusion and Future Work}
\label{sect:conclusion}


In this paper, we propose \our, an embedding framework to extract relation under heterogeneous supervision. 
When dealing with heterogeneous supervisions, one unique challenge is how to resolve conflicts generated by different labeling functions. 
Accordingly, we go beyond the ``source consistency assumption'' in prior works and leverage context-aware embeddings to induce proficient subsets.
The resulting framework bridges true label discovery and relation extraction with context representation, and allows them to mutually enhance each other. 
Experimental evaluation justifies the necessity of involving context-awareness, the quality of inferred true label, and the effectiveness of the proposed framework on two real-world datasets.

There exist several directions for future work. 
One is to apply transfer learning techniques to handle label distributions' difference between training set and test set. 
Another is to incorporate OpenIE methods to automatically find domain-specific patterns and generate pattern-based labeling functions.

%% file: acknowledge.tex
\section{Acknowledgments}
\label{sect:ack}

Research was sponsored in part by the U.S. Army Research Lab. under Cooperative Agreement No. W911NF-09-2-0053 (NSCTA), National Science Foundation IIS-1320617, IIS 16-18481, and NSF IIS 17-04532, and grant 1U54GM114838 awarded by NIGMS through funds provided by the trans-NIH Big Data to Knowledge (BD2K) initiative (www.bd2k.nih.gov). The views and conclusions contained in this document are those of the author(s) and should not be interpreted as representing the official policies of the U.S. Army Research Laboratory or the U.S. Government. The U.S. Government is authorized to reproduce and distribute reprints for Government purposes notwithstanding any copyright notation hereon.